\documentclass[10pt, conference, compsocconf]{IEEEtran}

\usepackage[utf8]{inputenc} 
\usepackage[T1]{fontenc}    
\usepackage{hyperref}       
\usepackage{url}            
\usepackage{booktabs}       
\usepackage{amsfonts}       
\usepackage{microtype}      
\usepackage{subfigure}
\usepackage{epsfig}
\usepackage{epstopdf}
\usepackage{graphicx}
\usepackage{amsmath}
\usepackage{amssymb}
\usepackage{subfigure}
\usepackage{float}

\begin{document}
%
\title{Tiny SSD: A Tiny Single-shot Detection Deep Convolutional Neural Network for Real-time Embedded Object Detection}


\author{\IEEEauthorblockN{Alexander Wong, Mohammad Javad Shafiee, Francis Li, Brendan Chwyl}
\IEEEauthorblockA{Dept. of Systems Design Engineering\\
	University of Waterloo, DarwinAI\\
	\{a28wong, mjshafiee\}@uwaterloo.ca, \{francis, brendan\}@darwinai.ca
}
}

\maketitle

\begin{abstract}
Object detection is a major challenge in computer vision, involving both object classification and object localization within a scene.  While deep neural networks have been shown in recent years to yield very powerful techniques for tackling the challenge of object detection, one of the biggest challenges with enabling such object detection networks for widespread deployment on embedded devices is high computational and memory requirements.  Recently, there has been an increasing focus in exploring small deep neural network architectures for object detection that are more suitable for embedded devices, such as Tiny YOLO and SqueezeDet.  Inspired by the efficiency of the Fire microarchitecture introduced in SqueezeNet and the object detection performance of the single-shot detection macroarchitecture introduced in SSD, this paper introduces Tiny SSD, a single-shot detection deep convolutional neural network for real-time embedded object detection that is composed of a highly optimized, non-uniform Fire sub-network stack and a non-uniform sub-network stack of highly optimized SSD-based auxiliary convolutional feature layers designed specifically to minimize model size while maintaining object detection performance.  The resulting Tiny SSD possess a model size of 2.3MB ($\sim$26X smaller than Tiny YOLO) while still achieving an mAP of 61.3\% on VOC 2007 ($\sim$4.2\% higher than Tiny YOLO).  These experimental results show that very small deep neural network architectures can be designed for real-time object detection that are well-suited for embedded scenarios.
\end{abstract}


\IEEEpeerreviewmaketitle

\section{Introduction}
\begin{figure}[t]
			\includegraphics[width = 4.2 cm]{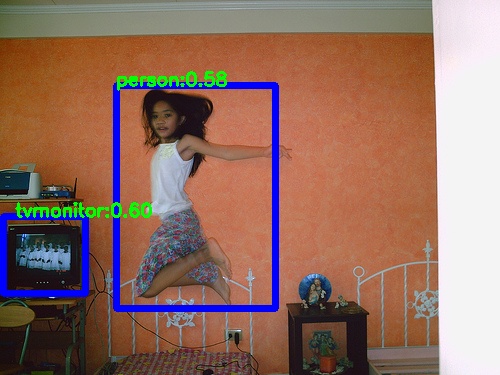}
			\includegraphics[width = 4.2 cm]{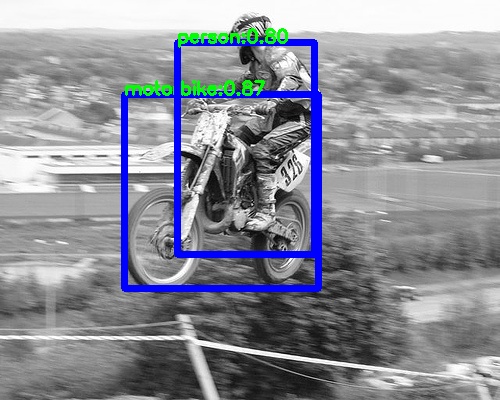}
		\caption{\textbf{Tiny SSD} results on the VOC test set.  The bounding boxes, categories, and confidences are shown.}
		\label{fig:example}
\end{figure}

Object detection can be considered a major challenge in computer vision, as it involves a combination of object classification and object localization within a scene (see Figure~\ref{fig:example}).  The advent of modern advances in deep learning~\cite{lecun2015deep,krizhevsky2012imagenet} has led to significant advances in object detection, with the majority of research focuses on designing increasingly more complex object detection networks for improved accuracy such as SSD~\cite{liu2016ssd}, R-CNN~\cite{girshick2014rich}, Mask R-CNN~\cite{MaskRCNN}, and other extended variants of these networks~\cite{huang2017speed,lin2017feature,shrivastava2016training}.  Despite the fact that such object detection networks have showed state-of-the-art object detection accuracies beyond what can be achieved by previous state-of-the-art methods, such networks are often intractable for use for embedded applications due to computational and memory constraints.  In fact, even faster variants of these networks such as Faster R-CNN~\cite{ren2015faster} are only capable of single-digit frame rates on a high-end graphics processing unit (GPU).  As such, more efficient deep neural networks for real-time embedded object detection is highly desired given the large number of operational scenarios that such networks would enable, ranging from smartphones to aerial drones.

Recently, there has been an increasing focus in exploring small deep neural network architectures for object detection that are more suitable for embedded devices.  For example, Redmon et al. introduced YOLO~\cite{redmon2016you} and YOLOv2~\cite{redmon2016yolo9000}, which were designed with speed in mind and was able to achieve real-time object detection performance on a high-end Nvidia Titan X desktop GPU.  However, the model size of YOLO and YOLOv2 remains very large in size (753 MB and 193 MB, respectively), making them too large from a memory perspective for most embedded devices.  Furthermore, their object detection speed drops considerably when running on embedded chips~\cite{shafiee2017fast}.  To address this issue, Tiny YOLO~\cite{tinyyolo} was introduced where the network architecture was reduced considerably to greatly reduce model size (60.5 MB) as well as greatly reduce the number of floating point operations required (just 6.97 billion operations) at a cost of object detection accuracy (57.1\% on the twenty-category VOC 2017 test set).  Similarly, Wu et al. introduced SqueezeDet~\cite{wu2016squeezedet}, a fully convolutional neural network that leveraged the efficient Fire microarchitecture introduced in SqueezeNet~\cite{iandola2016squeezenet} within an end-to-end object detection network architecture.  Given that the Fire microarchitecture is highly efficient, the resulting SqueezeDet had a reduced model size specifically for the purpose of autonomous driving.  However, SqueezeDet has only been demonstrated for objection detection with limited object categories (only three) and thus its ability to handle larger number of categories have not been demonstrated.  As such, the design of highly efficient deep neural network architectures that are well-suited for real-time embedded object detection while achieving improved object detection accuracy on a variety of object categories is still a challenge worth tackling.

In an effort to achieve a fine balance between object detection accuracy and real-time embedded requirements (i.e., small model size and real-time embedded inference speed), we take inspiration by both the incredible efficiency of the Fire microarchitecture introduced in SqueezeNet~\cite{iandola2016squeezenet} and the powerful object detection performance demonstrated by the single-shot detection macroarchitecture introduced in SSD~\cite{liu2016ssd}.  The resulting network architecture achieved in this paper is \textbf{Tiny SSD}, a single-shot detection deep convolutional neural network designed specifically for real-time embedded object detection.  Tiny SSD is composed of a non-uniform highly optimized Fire sub-network stack, which feeds into a non-uniform sub-network stack of highly optimized SSD-based auxiliary convolutional feature layers, designed specifically to minimize model size while retaining object detection performance.

This paper is organized as follows.  Section 2 describes the highly optimized Fire sub-network stack leveraged in the Tiny SSD network architecture.  Section 3 describes the highly optimized sub-network stack of SSD-based convolutional feature layers used in the Tiny SSD network architecture.  Section 4 presents experimental results that evaluate the efficacy of Tiny SSD for real-time embedded object detection.  Finally, conclusions are drawn in Section 5.

\section{Optimized Fire Sub-network Stack}

The overall network architecture of the Tiny SSD network for real-time embedded object detection is composed of two main sub-network stacks: i) a non-uniform Fire sub-network stack, and ii) a non-uniform sub-network stack of highly optimized SSD-based auxiliary convolutional feature layers, with the first sub-network stack feeding into the second sub-network stack.  In this section, let us first discuss in detail the design philosophy behind the first sub-network stack of the Tiny SSD network architecture: the optimized fire sub-network stack.

A powerful approach to designing smaller deep neural network architectures for embedded inference is to take a more principled approach and leverage architectural design strategies to achieve more efficient deep neural network microarchitectures~\cite{howard2017mobilenets,iandola2016squeezenet}.  A very illustrative example of such a principled approach is the SqueezeNet~\cite{iandola2016squeezenet} network architecture, where three key design strategies were leveraged:
\begin{enumerate}
\item reduce the number of $3\times3$ filters as much as possible,
\item reduce the number of input channels to $3\times3$ filters where possible, and
\item perform downsampling at a later stage in the network.
\end{enumerate}

\begin{figure}[!tp]
	\begin{center}
		\includegraphics[width = 8.6 cm]{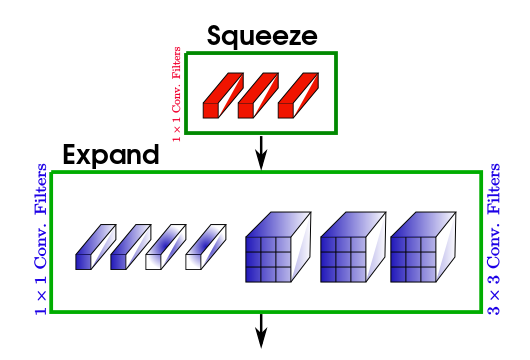}
	\end{center}
	\caption{An illustration of the Fire microarchitecture. The output of previous layer is squeezed by a \textit{squeeze} convolutional layer of $1\times1$ filters, which reduces the number of input channels to $3\times3$ filters.  The result of the squeeze convolutional layers is passed into the \textit{expand} convolutional layer which consists of both $1\times1$ and $3\times3$ filters.      }
	\label{fig:fire-diagram}
\end{figure}

This principled designed strategy led to the design of what the authors referred to as the \textbf{Fire} module, which consists of a \textit{squeeze} convolutional layer of 1x1 filters (which realizes the second design strategy of effectively reduces the number of input channels to $3\times3$ filters) that feeds into an \textit{expand} convolutional layer comprised of both $1\times1$ filters and $3\times3$ filters (which realizes the first design strategy of effectively reducing the number of $3\times3$ filters).  An illustration of the Fire microarchitecture is shown in Figure~\ref{fig:fire-diagram}.

Inspired by the elegance and simplicity of the Fire microarchitecture design, we design the first sub-network stack of the Tiny SSD network architecture as a standard convolutional layer followed by a set of highly optimized Fire modules.  One of the key challenges to designing this sub-network stack is to determine the ideal number of Fire modules as well as the ideal microarchitecture of each of the Fire modules to achieve a fine balance between object detection performance and model size as well as inference speed.  First, it was determined empirically that 10 Fire modules in the optimized Fire sub-network stack provided strong object detection performance. In terms of the ideal microarchitecture, the key design parameters of the Fire microarchitecture are the number of filters of each size ($1\times1$ or $3\times3$) that form this microarchitecture.  In the SqueezeNet network architecture that first introduced the Fire microarchitecture~\cite{iandola2016squeezenet}, the microarchitectures of the Fire modules are largely uniform, with many of the modules sharing the same microarchitecture configuration.  In an effort to achieve more optimized Fire microarchitectures on a per-module basis, the number of filters of each size in each Fire module is optimized to have as few parameters as possible while still maintaining the overall object detection accuracy.  As a result, the optimized Fire sub-network stack in the Tiny SSD network architecture is highly non-uniform in nature for an optimal sub-network architecture configuration.  Table~\ref{tab:TinySDDFirstPart} shows the overall architecture of the highly optimized Fire sub-network stack in Tiny SSD, and the number of parameters in each layer of the sub-network stack.

\begin{figure}[!tp]
	\begin{center}
		\includegraphics[width = 8.6 cm]{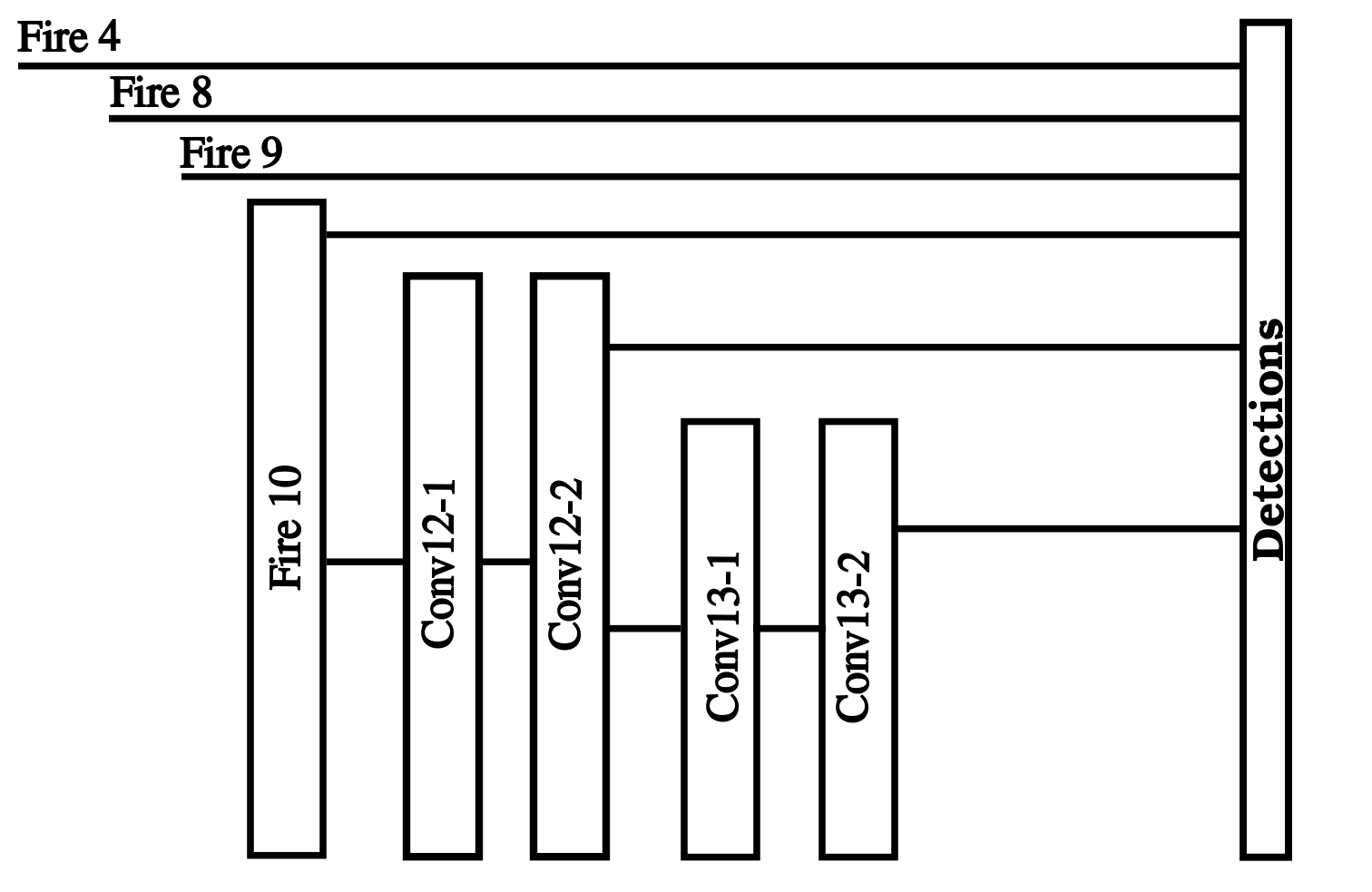}
	\end{center}
	\caption{An illustration of the network architecture of the second sub-network stack of Tiny SSD.  The output of three Fire modules and two auxiliary convolutional feature layers, all with highly optimized microarchitecture configurations, are combined together for object detection.}
	\label{fig:SSD-diagram}
\end{figure}

\section{Optimized Sub-network Stack of SSD-based Convolutional Feature Layers}

In this section, let us first discuss in detail the design philosophy behind the second sub-network stack of the Tiny SSD network architecture: the sub-network stack of highly optimized SSD-based auxiliary convolutional feature layers.

\begin{table}
	\begin{center}
		\small
		\caption{The optimized Fire sub-network stack of the Tiny SSD network architecture. The number of filters and input size to each layer are reported for the convolutional layers and Fire modules. Each Fire module is reported in one row for a better representation. ''$x$@S -- $y$@E1 -- $z$@E3" stands for $x$ numbers of $1\times1$ filters in the squeeze convolutional layer, $y$ numbers of $1\times1$ filters and $z$ numbers of $3\times3$ filters in the expand convolutional layer. }
		\label{tab:TinySDDFirstPart}
		\begin{tabular}{c|c|c}
			\hline \hline
			Type / Stride & Filter Shapes & Input Size \\ \hline
			Conv1 / s2& $3\times3\times57$ &$300\times300$   \\ \hline
			Pool1 / s2 & $3 \times 3$ & $149\times149$\\ \hline
			Fire1&  15@S -- 49@E1 -- 53@E3& $74\times 74$ \\ \hline
			\multicolumn{3}{c}{Concat1}\\ \hline
			Fire2& 15@S -- 54@E1 -- 52@E3 & $74\times 74$\\ \hline
			\multicolumn{3}{c}{Concat2}\\\hline
			Pool3 / s2 & $3 \times 3$ & $74\times 74$ \\ \hline
			Fire3&29@S -- 92@E1 -- 94@E3  &$37\times 37$ \\ \hline
			\multicolumn{3}{c}{Concat3}\\\hline
			Fire4& 29@S -- 90@E1 -- 83@E3 &$37\times 37$ \\ \hline
			\multicolumn{3}{c}{Concat4}\\\hline
			Pool5 / s2 & $3 \times 3$ &$37\times 37$  \\ \hline
			Fire5& 44@S -- 166@E1 -- 161@E3 &$18\times 18$ \\ \hline
			\multicolumn{3}{c}{Concat5}\\\hline
			Fire6&  45@S -- 155@E1 -- 146@E3&$18\times 18$ \\ \hline
			\multicolumn{3}{c}{Concat6}\\\hline
			Fire7& 49@S -- 163@E1 -- 171@E3 &$18\times 18$ \\ \hline
			\multicolumn{3}{c}{Concat7}\\\hline
			Fire8&  25@S -- 29@E1 -- 54@E3& $18\times 18$\\ \hline
			\multicolumn{3}{c}{Concat8}\\\hline
			Pool9 / s2 & $3 \times 3$ &$18\times 18$\\ \hline
			Fire 9&  37@S -- 45@E1 -- 56@E3&$9\times 9$ \\ \hline
			\multicolumn{3}{c}{Concat9}\\\hline
			Pool10 / s2 & $3 \times 3$ & \\ \hline
			Fire10&38@S -- 41@E1 -- 44@E3  & $4\times 4$\\ \hline
			\multicolumn{3}{c}{Concat10}\\\hline
		\end{tabular}
		\vspace{-0.6cm}
	\end{center}
\end{table}

One of the most widely-used and effective object detection network macroarchitectures in recent years has been the single-shot multibox detection (SSD) macroarchitecture~\cite{liu2016ssd}.  The SSD macroarchitecture augments a base feature extraction network architecture with a set of auxiliary convolutional feature layers and convolutional predictors.  The auxiliary convolutional feature layers are designed such that they decrease in size in a progressive manner, thus enabling the flexibility of detecting objects within a scene across different scales.  Each of the auxiliary convolutional feature layers can then be leveraged to obtain either: i) a confidence score for a object category, or ii) a shape offset relative to default bounding box coordinates~\cite{liu2016ssd}.  As a result, a number of object detections can be obtained per object category in this manner in a powerful, end-to-end single-shot manner.

Inspired by the powerful object detection performance and multi-scale flexibility of the SSD macroarchitecture~\cite{liu2016ssd}, the second sub-network stack of Tiny SSD is comprised of a set of auxiliary convolutional feature layers and convolutional predictors with highly optimized microarchitecture configurations (see Figure~\ref{fig:SSD-diagram}).

As with the Fire microarchitecture, a key challenge to designing this sub-network stack is to determine the ideal microarchitecture of each of the auxiliary convolutional feature layers and convolutional predictors to achieve a fine balance between object detection performance and model size as well as inference speed.  The key design parameters of the auxiliary convolutional feature layer microarchitecture are the number of filters that form this microarchitecture.  As such, similar to the strategy taken for constructing the highly optimized Fire sub-network stack, the number of filters in each auxiliary convolutional feature layer is optimized to minimize the number of parameters while preserving overall object detection accuracy of the full Tiny SSD network. As a result, the optimized sub-network stack of auxiliary convolutional feature layers in the Tiny SSD network architecture is highly non-uniform in nature for an optimal sub-network architecture configuration.  Table~\ref{tab:ssdsecondpart} shows the overall architecture of the optimized sub-network stack of the auxiliary convolutional feature layers within the Tiny SSD network architecture, along with the number of parameters in each layer.

\begin{figure*}[t]
	\begin{center}
		\begin{tabular}{cccc}
			\includegraphics[width = 4 cm]{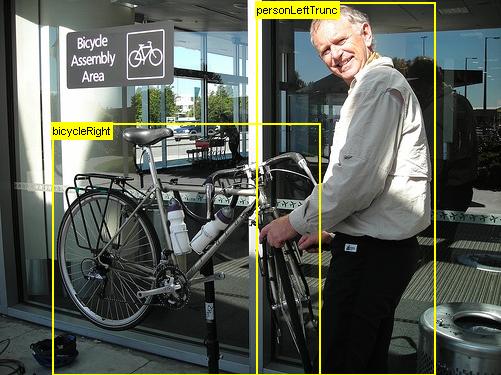}&
			\includegraphics[width = 4 cm]{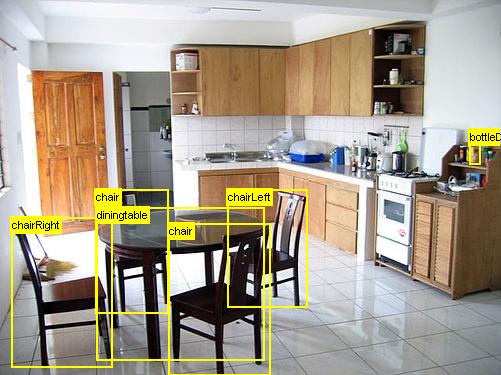}&
			\includegraphics[width = 4 cm]{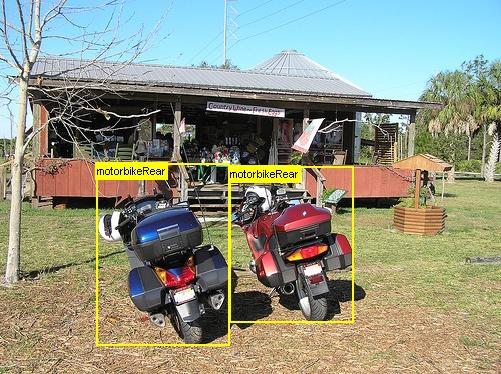}&
			\includegraphics[width = 4 cm]{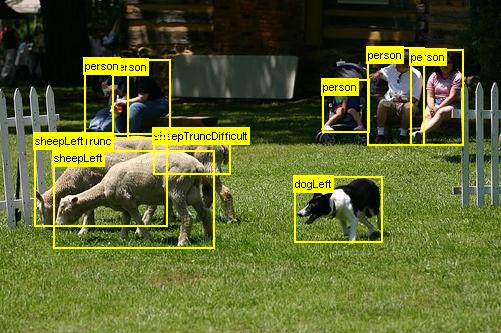}
		\end{tabular}
		\caption{ Example images from the Pascal VOC dataset.  The ground-truth bounding boxes and object categories are shown for each image.   }
		\label{fig:sample}
	\end{center}
\end{figure*}

\begin{table}
	\begin{center}
		\small
		\caption{The optimized sub-network stack of the auxiliary convolutional feature layers within the Tiny SSD network architecture.  The input sizes to each convolutional layer and kernel sizes are reported.}
		\label{tab:ssdsecondpart}
		\begin{tabular}{l|l|c}
			\hline \hline
			Type / Stride     & Filter Shape       & Input Size \\ \hline
			Conv12-1 / s2     & $ 3\times3 \times 51$  & $4 \times 4$\\ \hline
			Conv12-2          & $ 3\times3 \times 46$  &  $4 \times 4$\\ \hline
			Conv13-1          & $ 3\times3 \times 55 $  &  $2 \times 2$\\ \hline
			Conv13-2          & $ 3\times3 \times 85$  &  $2 \times 2$\\ \hline
			Fire5-mbox-loc    & $ 3\times3 \times16 $  & $37 \times 37$\\ \hline
			Fire5-mbox-conf   &$ 3\times 3\times84 $   &  $37 \times 37$\\ \hline
			Fire9-mbox-loc    &$ 3\times3 \times24 $   &  $18 \times 18$\\ \hline
			Fire9-mbox-conf   &$ 3\times3 \times 126$   & $18 \times 18$ \\ \hline
			Fire10-mbox-loc   & $ 3\times3 \times 24$  &  $9 \times 9$\\ \hline
			Fire10-mbox-conf  &$ 3\times3 \times126 $   &  $9 \times 9$\\ \hline
			Fire11-mbox-loc   & $ 3\times3 \times24 $  &  $4 \times 4$\\ \hline
			Fire11-mbox-conf  &$ 3\times3 \times126 $   & $4 \times 4$\\ \hline
			Conv12-2-mbox-loc & $ 3\times3 \times24 $  & $2 \times 2$\\ \hline
			Conv12-2-mbox-conf&$ 3\times3 \times 126$   & $2 \times 2$\\ \hline
			Conv13-2-mbox-loc &$ 3\times3 \times16 $   &  $1 \times 1$\\ \hline
			Conv13-2-mbox-conf& $ 3\times3 \times84 $  &  $1 \times 1$\\ \hline

		\end{tabular}
	\end{center}
\end{table}

\begin{table}[ht]
	\begin{center}
		\small
		\caption{Object detection accuracy results of Tiny SSD on VOC 2007 test set.  Tiny YOLO results are provided as a baseline comparison. }
		\label{Tab:res}
		\begin{tabular}{|c|c||c|c|c|c|}
			\hline
			Model  & Model  & mAP  \\
			Name & size  & (VOC 2007)   \\\hline \hline
			Tiny YOLO~\cite{tinyyolo} & 60.5MB &57.1\% \\
			Tiny SSD & 2.3MB & 61.3\% \\\hline
		\end{tabular}
	\end{center}
\end{table}

\begin{table}[ht]
	\begin{center}
		\small
		\caption{Resource usage of Tiny SSD.}
		\label{Tab:macs-runtime}
		\begin{tabular}{|c|c|c|}
			\hline
			Model  &Total number & Total number \\
			Name &of Parameters &of MACs    \\\hline \hline
			Tiny SSD &1.13M & 571.09M  \\\hline
		\end{tabular}
	\end{center}
\end{table}

\section{Parameter Precision Optimization}
In this section, let us discuss the parameter precision optimization strategy for Tiny SSD.  For embedded scenarios where the computational requirements and memory requirements are more strict, an effective strategy for reducing computational and memory footprint of deep neural networks is reducing the data precision of parameters in a deep neural network.  In particular, modern CPUs and GPUs have moved towards accelerated mixed precision operations as well as better handling of reduced parameter precision, and thus the ability to take advantage of these factors can yield noticeable improvements for embedded scenarios.  For Tiny SSD, the parameters are represented in half precision floating-point, thus leading to further deep neural network model size reductions while having a negligible effect on object detection accuracy.

\section{Experimental Results and Discussion}
\vspace{-0.1 cm}
To study the utility of Tiny SSD for real-time embedded object detection, we examine the model size, object detection accuracies, and computational operations on the VOC2007/2012 datasets.  For evaluation purposes, the Tiny YOLO network~\cite{tinyyolo} was used as a baseline reference comparison given its popularity for embedded object detection, and was also demonstrated to possess one of the smallest model sizes in literature for object detection on the VOC 2007/2012 datasets (only 60.5MB in size and requiring just 6.97 billion operations).  The VOC2007/2012 datasets consist of natural images that have been annotated with 20 different types of objects, with illustrative examples shown in Figure~\ref{fig:sample}.  The tested deep neural networks were trained using the VOC2007/2012 training datasets, and the mean average precision (mAP) was computed on the VOC2007 test dataset to evaluate the object detection accuracy of the deep neural networks.

\subsection{Training Setup}
The proposed Tiny SSD network was trained for 220,000 iterations in the Caffe framework with training batch size of 24.  RMSProp was utilized as the training policy with base learning rate set to 0.00001 and $\gamma = 0.5$.

\begin{figure*}
	\begin{center}
		\begin{tabular}{ccc}
			
			\includegraphics[width = 4.3 cm]{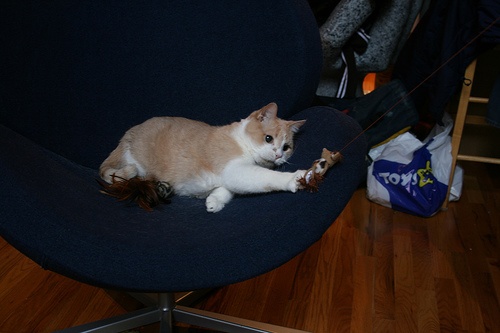}&
			\includegraphics[width = 4.3 cm]{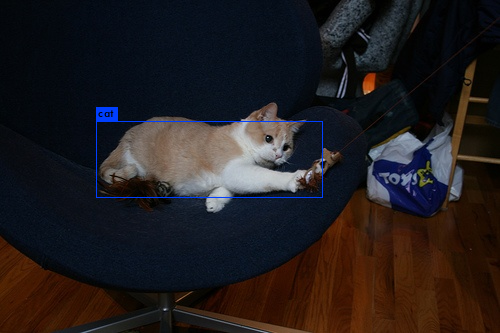}&
			\includegraphics[width = 4.3 cm]{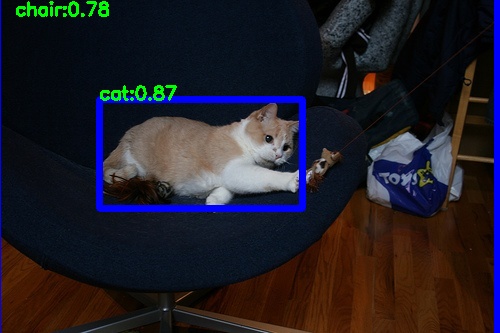}  \\
			\includegraphics[width = 4.3 cm]{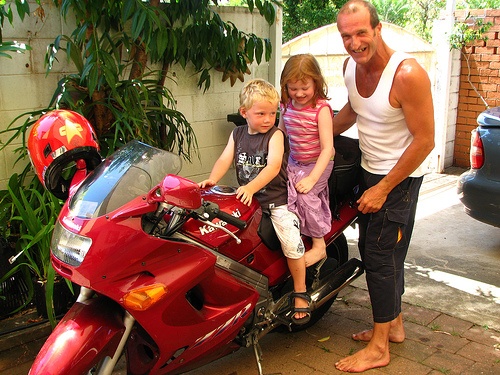}&
			\includegraphics[width = 4.3 cm]{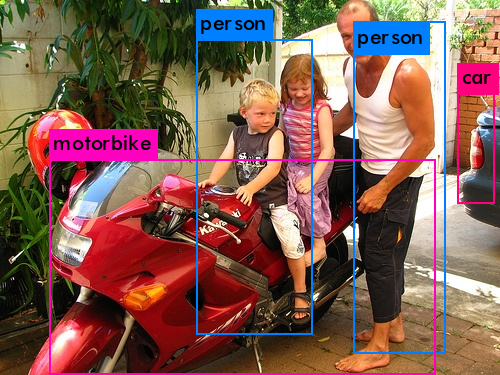}&
			\includegraphics[width = 4.3 cm]{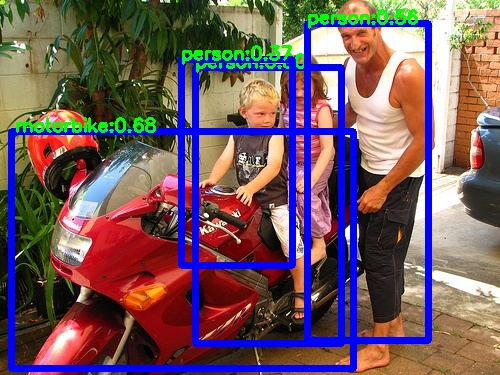}  \\
			\includegraphics[width = 4.3 cm]{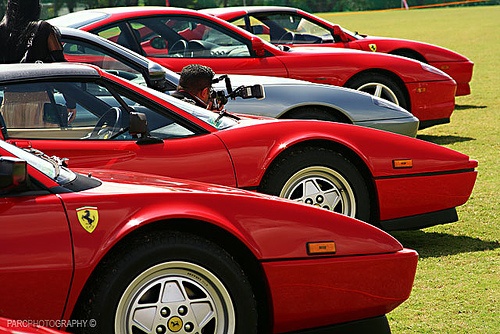}&
			\includegraphics[width = 4.3 cm]{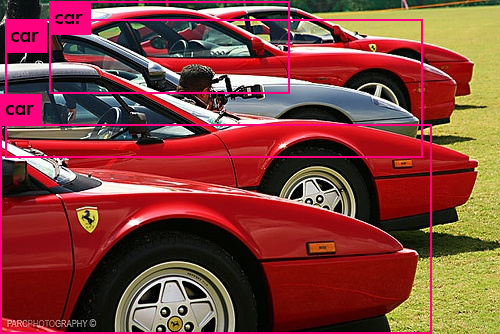}&
			\includegraphics[width = 4.3 cm]{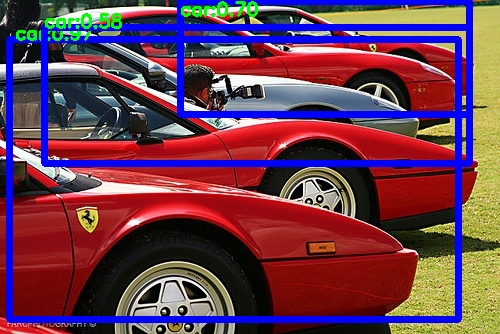}  \\
			\includegraphics[width = 4.3 cm]{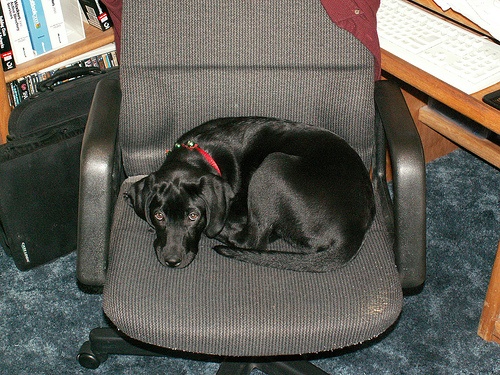}&
			\includegraphics[width = 4.3 cm]{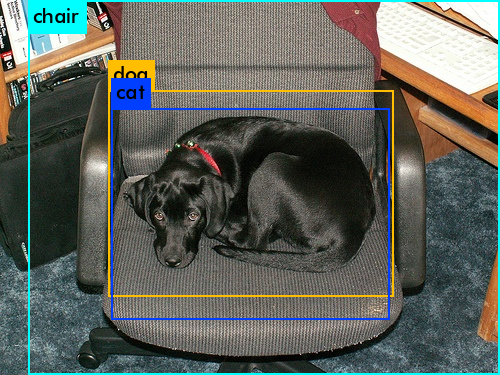}&
			\includegraphics[width = 4.3 cm]{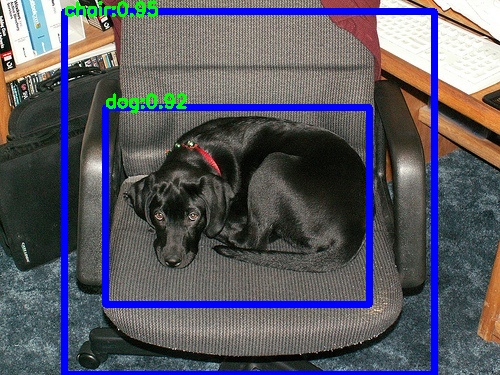}  \\
			\includegraphics[width = 4.3 cm]{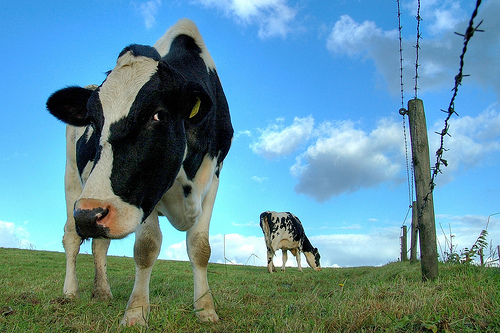}&
			\includegraphics[width = 4.3 cm]{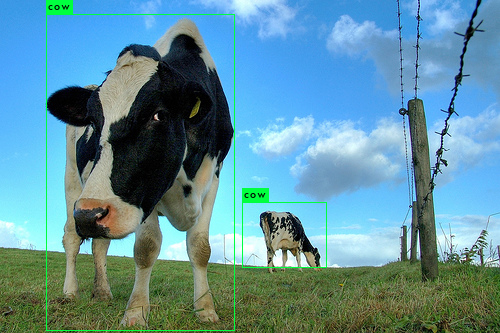}&
			\includegraphics[width = 4.3 cm]{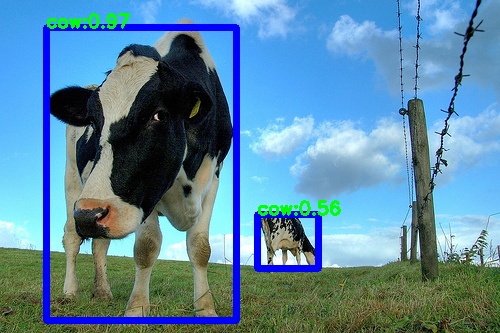}  \\
			\includegraphics[width = 4.3 cm]{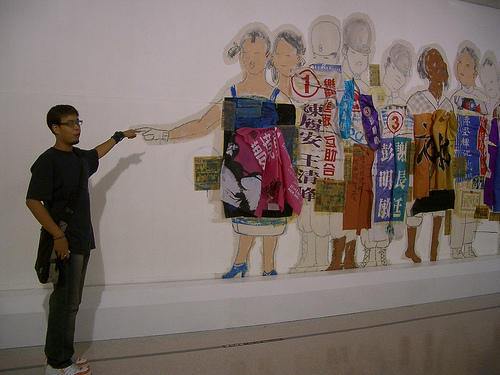}&
			\includegraphics[width = 4.3 cm]{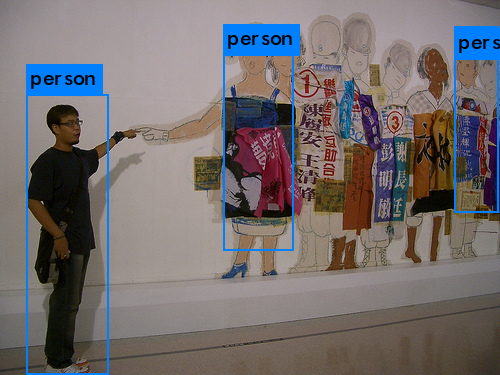}&
			\includegraphics[width = 4.3 cm]{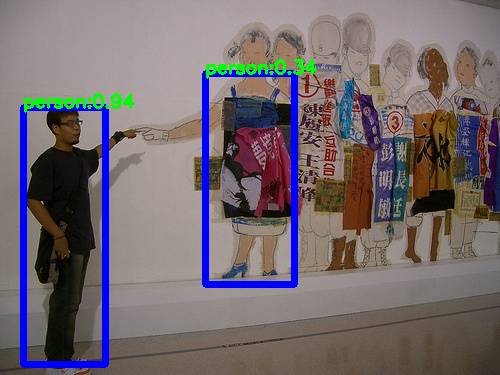}  \\
			Input Image & Tiny YOLO& Tiny SSD
		\end{tabular}
		\caption{ Example object detection results produced by the proposed Tiny SSD compared to Tiny YOLO. It can be observed that Tiny SSD has comparable object detection results as Tiny YOLO in some cases, while in some cases outperforms Tiny YOLO in assigning more accurate category labels to detected objects.  This significant improvement in object detection accuracy when compared to Tiny YOLO illustrates the efficacy of Tiny SSD for providing more reliable embedded object detection performance.  }
		\label{fig:res}
	\end{center}
\end{figure*}

\subsection{Discussion}
Table~\ref{Tab:res} shows the model size and the object detection accuracy of the proposed Tiny SSD network on the VOC 2007 test dataset, along with the model size and the object detection accuracy of Tiny YOLO.  A number of interesting observations can be made.  First, the resulting Tiny SSD possesses a model size of 2.3MB, which is $\sim$26X smaller than Tiny YOLO.  The significantly smaller model size of Tiny SSD compared to Tiny YOLO illustrates its efficacy for greatly reducing the memory requirements for leveraging Tiny SSD for real-time embedded object detection purposes.  Second, it can be observed that the resulting Tiny SSD was still able to achieve an mAP of 61.3\% on the VOC 2007 test dataset, which is $\sim$4.2\% higher than that achieved using Tiny YOLO.  Figure~~\ref{fig:res} demonstrates several example object detection results produced by the proposed Tiny SSD compared to Tiny YOLO. It can be observed that Tiny SSD has comparable object detection results as Tiny YOLO in some cases, while in some cases outperforms Tiny YOLO in assigning more accurate category labels to detected objects.  For example, in the first image case, Tiny SSD is able to detect the chair in the scene, while Tiny YOLO misses the chair.  In the third image case, Tiny SSD is able to identify the dog in the scene while Tiny YOLO detects two bounding boxes around the dog, with one of the bounding boxes incorrectly labeling it as cat.  This significant improvement in object detection accuracy when compared to Tiny YOLO illustrates the efficacy of Tiny SSD for providing more reliable embedded object detection performance.   Furthermore, as seen in Table~\ref{Tab:macs-runtime}, Tiny SSD requires just 571.09 million MAC operations to perform inference, making it well-suited for real-time embedded object detection.  These experimental results show that very small deep neural network architectures can be designed for real-time object detection that are well-suited for embedded scenarios.

\section{Conclusions}
In this paper, a single-shot detection deep convolutional neural network called Tiny SSD is introduced for real-time embedded object detection.  Composed of a highly optimized, non-uniform Fire sub-network stack and a non-uniform sub-network stack of highly optimized SSD-based auxiliary convolutional feature layers designed specifically to minimize model size while maintaining object detection performance, Tiny SSD possesses a model size that is $\sim$26X smaller than Tiny YOLO, requires just 571.09 million MAC operations, while still achieving an mAP of that is $\sim$4.2\% higher than Tiny YOLO on the VOC 2007 test dataset.  These results demonstrates the efficacy of designing very small deep neural network architectures such as Tiny SSD for real-time object detection in embedded scenarios.

\section*{Acknowledgment}

The authors thank Natural Sciences and Engineering Research Council of Canada, Canada Research Chairs Program, DarwinAI, and Nvidia for hardware support.

{\small
	\bibliographystyle{plain}
	\bibliography{ccn_style}
}
\end{document}